\documentclass[letterpaper]{article} 
\usepackage{aaai24}  
\usepackage{times}  
\usepackage{helvet}  
\usepackage{courier}  
\usepackage[hyphens]{url}  
\usepackage{graphicx} 
\usepackage{natbib}  
\usepackage{caption} 
\usepackage{algorithm}
\usepackage{algorithmic}

\usepackage{amsfonts}
\usepackage{amsmath}
\usepackage{bm}
\usepackage{booktabs}
\usepackage{dsfont}
\usepackage{multirow}
\usepackage{color, colortbl}
\definecolor{Gray}{gray}{0.9}
\usepackage{graphicx}
\usepackage{subfigure}
\usepackage{caption}
\usepackage{makecell}

\usepackage{newfloat}
\usepackage{listings}
\DeclareCaptionStyle{ruled}{labelfont=normalfont,labelsep=colon,strut=off} 
\floatstyle{ruled}
\newfloat{listing}{tb}{lst}{}
\floatname{listing}{Listing}
\title{Controller-Guided Partial Label Consistency Regularization with Unlabeled Data}
\author{
    Qian-Wei Wang\textsuperscript{\rm 1,\rm 2},
    Bowen Zhao\textsuperscript{\rm 1},
    Mingyan Zhu\textsuperscript{\rm 1},
    Tianxiang Li\textsuperscript{\rm 1},
    Zimo Liu\textsuperscript{\rm 2},
    Shu-Tao Xia\textsuperscript{\rm 1, \rm 2}\thanks{Corresponding author.}
}
\affiliations{
    \textsuperscript{\rm 1}Tsinghua Shenzhen International Graduate School, Tsinghua University\\
    \textsuperscript{\rm 2}Research Center of Artificial Intelligence, Peng Cheng Laboratory\\
    \{wanggw21, zbm18, zmy20, litx21\}@mails.tsinghua.edu.cn, liuzm@pcl.ac.cn, xiast@sz.tsinghua.edu.cn
}

\begin{document}

\maketitle

\begin{abstract}
Partial label learning (PLL) learns from training examples each associated with multiple candidate labels, among which only one is valid. In recent years, benefiting from the strong capability of dealing with ambiguous supervision and the impetus of modern data augmentation methods, consistency regularization-based PLL methods have achieved a series of successes and become mainstream. However, as the partial annotation becomes insufficient, their performances drop significantly. In this paper, we leverage easily accessible unlabeled examples to facilitate the partial label consistency regularization. In addition to a partial supervised loss, our method performs a controller-guided consistency regularization at both the label-level and representation-level with the help of unlabeled data. To minimize the disadvantages of insufficient capabilities of the initial supervised model, we use the controller to estimate the confidence of each current prediction to guide the subsequent consistency regularization. Furthermore, we dynamically adjust the confidence thresholds so that the number of samples of each class participating in consistency regularization remains roughly equal to alleviate the problem of class-imbalance. Experiments show that our method achieves satisfactory performances in more practical situations, and its modules can be applied to existing PLL methods to enhance their capabilities.
\end{abstract}

\section{Introduction}

In real-world applications, data with unique and correct label is often too costly to obtain \cite{zhou2018brief, li2019towards, wang2020learning, liu2023consistent, chen2023imprecise}. Instead, users with varying knowledge and cultural backgrounds tend to annotate the same image with different labels. Traditional supervised learning framework base on "one instance one label" assumption is out of its capability when faced with such ambiguous examples, while partial label learning (PLL) provides an effective solution. Conceptually speaking, PLL \cite{Hullermeier_Beringer2006, Nguyen_Caruana2008, Cour_2011, ZY15, Yu_Zhang2017, DBLP:conf/nips/FengL0X0G0S20, DBLP:journals/tkde/LyuFWLL21, DBLP:conf/icml/LvXF0GS20, DBLP:conf/icml/WenCHL0L21, wang2023deep, shi2023robust} learns from ambiguous labeling information where each training example is associated with a set of candidate labels, among which only one is assumed to be valid. The key to accomplish this task is to find the correct correspondence between each training example and its ground-truth label from the ambiguous candidate collections, i.e. disambiguation. In recent years, partial label learning manifested its capability in solving a wide range of applications such as multimedia contents analysis \cite{Zeng_2013, DBLP:journals/pami/ChenPC18, xie2018partial}, web mining \cite{Jie_Orabona2010}, ecoinformatics \cite{Liu_Dietterich2012}, etc.

\begin{figure}[t]
\centering
\includegraphics[scale=0.45]{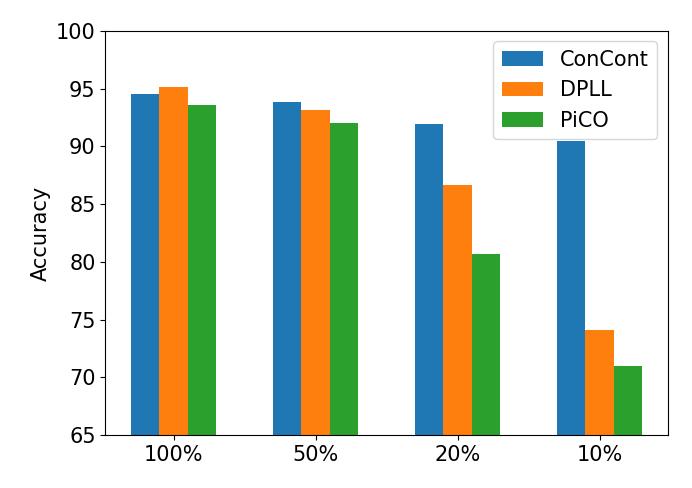}
\caption{The performances of ConCont (with additional unlabeled examples), DPLL and PiCO with $100\%$, $50\%$, $20\%$ and $10\%$ partially annotated examples on CIFAR-10 dataset.}
\label{example}
\end{figure}

With the explosive research of PLL in the deep learning paradigm, consistency regularized disambiguation-based methods \cite{wang2022pico, wu2022revisiting, wang2022pico+, li2023learning, xia2023towards} have achieved significantly better results than other solutions, and have gradually become the mainstream. Such methods usually perturb the samples in the feature space without changing their label semantics, and then use various methods in the label space or representation space to make the outputs of different variants consistent. Modern data augmentation methods \cite{cubuk2019autoaugment, DBLP:journals/corr/abs-1909-13719, devries2017improved} further improve their performances.

However, their performances are achieved when the number of partially labeled examples is sufficient. In real-world applications, PLL is usually in a scenario where the labeling resources are constrained, and the adequacy of partial annotation is not guaranteed. Under this circumstance, existing consistency regularization-based methods often fails to achieve satisfactory performances. As shown in Figure \ref{example}, DPLL \cite{wu2022revisiting} and PiCO \cite{wang2022pico} achieve state-of-the-art performances when using complete partial training set, but as the proportion of partial example decreases, their accuracies drop significantly. The reason behind this phenomenon is that when the number of labels is scarce and inherently ambiguous, there is not enough supervision information to guide the initial supervised learning of the model, which leads to the convergence of the consistency regularization to the wrong direction and the emergence of problems such as overfitting and class-imbalance.

Witnessing the enormous power of unlabeled examples via consistency regularization \cite{DBLP:conf/nips/SohnBCZZRCKL20, berthelot2019remixmatch, xie2020unsupervised}, we hope to facilitate partial label consistency regularization through these readily available data. To this end, an effective framework needs to be designed to maximize the potential of partial and unlabeled examples, as well as reasonable mechanisms to guide the model when supervision information is scarce and ambiguous. In this paper, we propose consistency regularization with controller (abbreviated as ConCont). Our method learns from the supervised information in the training targets (i.e., candidate label sets) via a supervised loss, while performing controller-guided consistency regularization at both label- and representation-levels with the help of unlabeled data. To avoid negative regularization, the controller divides the examples as confident or unconfident according to the prior information and the learning state of the model, and applies different label- and representation-level consistency regularization strategies, respectively. Furthermore, we dynamically adjust the confidence thresholds so that the number of samples of each class participating in consistency regularization remains roughly equal to alleviate the problem of class-imbalance.

\section{Related Work}

Traditional PLL methods can be divided into two categories: averaging-based \cite{Hullermeier_Beringer2006, Cour_2011, zhang2016partial} and identification-based \cite{DBLP:conf/nips/JinG02, Liu_Dietterich2012, DBLP:conf/aaai/FengA19, ni2021partial}. Averaging-based methods treat all the candidate labels equally, while identification-based methods aim at identifying the ground-truth label directly from candidate label set. With the popularity of deep neural networks, PLL has been increasingly studied in deep learning paradigm. Yao et al. \shortcite{yao2020deep} made the first attempt with an entropy-based regularizer enhancing discrimination. Lv et al. \shortcite{DBLP:conf/icml/LvXF0GS20} propose a classifier-consistent risk estimator for partial examples that theoretically converges to the optimal point learned from its fully supervised counterpart under mild condition, as well as an effective method progressively identifying ground-truth labels from the candidate sets. Wen et al. \shortcite{DBLP:conf/icml/WenCHL0L21} propose a family of loss functions named leveraged weighted loss taking the trade-offs between losses on partial labels and non-partial labels into consideration, advancing the former method to a more generalized case. Xu et al. \shortcite{xu2021instance} consider the learning case where the candidate labels are generated in an instance-dependent manner.

Recently, consistency regularization-based PLL methods have achieved impressive results, among which two representatives are: PiCO \cite{wang2022pico} and DPLL \cite{wu2022revisiting}. They can be seen as performing consistency regularization at the representation-level and label-level, respectively. To be specific, PiCO aligns the representations of the augmented variants of samples belonging to the same class, and calculates a representation prototype for each class, and then disambiguates the label distribution of the sample according to the distance between the sample representation and each class prototype, forming an iterative EM-like optimization. While DPLL aligns the output label distributions of multiple augmented variants to a conformal distribution, which serves as a comprehensiveness of the label distribution for all augmentations. Despite achieving state-of-the-art performances under fully PLL datasets, their consistency regularization rely heavily on the sufficiency of partial annotations, which greatly limits their applications.

Our work is also related with semi-supervised PLL \cite{wang2019partial, wang2020semi}. Despite similar learning scenarios, previous semi-supervised PLL methods are all based on nearest-neighbor or linear classifiers, and have not been integrated with modern consistency regularized deep learning, which are very different with our method in terms of algorithm implementation.

\section{Methodology}
\subsection{Notations}

Let $\mathcal{X} \subseteq \mathbb{R}^d$ be the input feature space and $\mathcal{Y} = \lbrace 1,2, \dots, C \rbrace$ denote the label space. We attempt to induce a multi-class classifier $f: \mathcal{X} \mapsto [0,1]^C$ from partial label training set $\mathcal{D}_p = \lbrace (\bm x^i , \bm y^i) | 1 \leq i \leq p \rbrace$ and an additional unlabeled set $\mathcal{D}_u = \lbrace \bm x^i | p+1 \leq i \leq p+u \rbrace$. Here, $\bm y^i = (y^i_1, y^i_2, \dots, y^i_C) \in \{0, 1\}^C$ is the partial label, in which $y^i_j = 1$ indicates the $j$-th label belonging to the candidate set of sample $\bm x^i$ and vice versa. Following the basic assumption of PLL, the ground-truth label $\ell \in \mathcal{Y}$ is inaccessible to the model while satisfies $y^i_{\ell} = 1$. In order to facilitate the unified operation of partial and unlabeled examples, the target vector of unlabeled examples is represented as $\bm (1, 1, \dots, 1)$, i.e., the candidate set equals to $\mathcal{Y}$, containing no label information. For the classifier $f$, we use $f_j(\bm x)$ to denote the output of classifier $f$ on label $j$ given input $\bm x$.

\subsection{Consistency Regularization with Controller}

Briefly, our method learns from the supervised information in the training targets (i.e., candidate label sets) via a supervised loss, while performing controller-guided consistency regularization at both label- and representation-level with the help of unlabeled data. The consistency regularization here works by aligning the outputs of different augmented variants of each example. To prevent the model from falling into poor convergence due to ambiguity and sparsity of supervision, we design a controller to divide the examples according to the prior information and the learning state of the model to guide the subsequent consistency regularization.

\begin{figure}[t]
\centering
\includegraphics[scale=0.20]{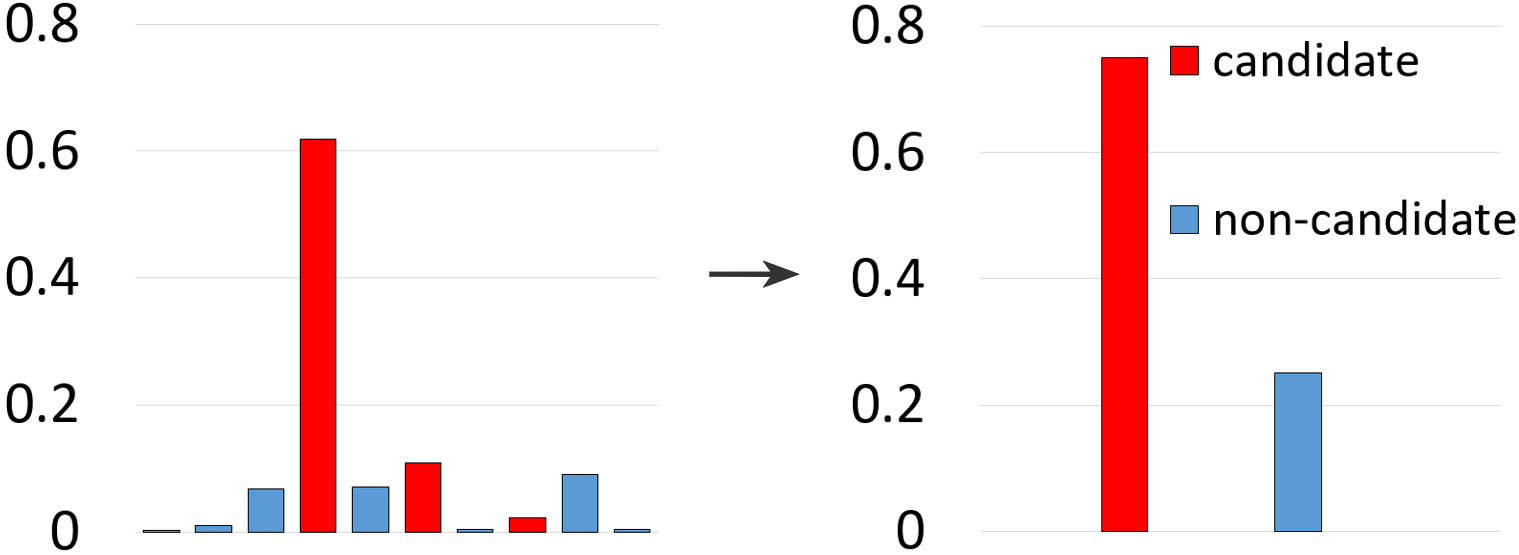}
\caption{Partial cross-entropy loss transforms multi-class predicted distribution into binary.}
\label{classification}
\end{figure}

Due to the inaccessibility of unique-label, it is infeasible to train the neural network by minimizing the cross-entropy loss between model prediction and training target as usual. To overcome this difficulty, the original multi-class classification is transformed into a binary classification in our method according to the implication of partial label, i.e. the input should belong to one of the candidate classes. In detail, our method aggregates the predicted probabilities over candidate labels and non-candidate labels respectively (see Figure \ref{classification}), forming a binary distribution over class \emph{candidate} and class \emph{non-candidate}. Then, the instance is classified to class \emph{candidate} with a simple binary cross-entropy loss.

Given instance $\bm x$ and its training target $\bm y$, the partial cross-entropy loss can be expressed as:
\begin{equation}
\mathcal{L}_{\textrm{part}} = - \log{\sum_{j=1}^{C} y_j f_j(\bm x)}.
\end{equation}

We encourage the model's output to remain invariant to perturbations applied to input images that do not change the label semantics. With the guidance of controller, our method minimizes the divergence of the output from a pair of different data augmentations: $(\textrm{Aug}_w(\cdot), \textrm{Aug}_s(\cdot))$ at the label-level and $(\textrm{Aug}_s(\cdot), \textrm{Aug}_s'(\cdot))$ at the representation-level (shown in Figure \ref{main}), where $\textrm{Aug}_w(\cdot)$ denotes the weak augmentation, $\textrm{Aug}_s(\cdot)$ and $\textrm{Aug}_s'(\cdot)$ denote two different strong augmentations. The reason for this design is that the predicted label distribution from the weakly-augmented variant is more reliable, which is then used to generate the pseudo-label of the input example for subsequent consistency regularization; while previous studies \cite{DBLP:conf/cvpr/He0WXG20, chen2020simple} have shown that the representation alignment of more diverse augmented variants is more conductive to improve the model's ability. Moreover, we also explored the "two-branch" framework, i.e., performs label- and representation-level consistency regularization on the same pair of augmentations $(\textrm{Aug}_w(\cdot), \textrm{Aug}_s(\cdot))$. However, the representation learning of this approach is less effective, meanwhile, sharing the same pair of augmentations may lead to overfitting. Experiments also confirmed that its performance is not as good as the adopted "three-branch" version.

\begin{figure}[t]
\centering
\includegraphics[scale=0.25]{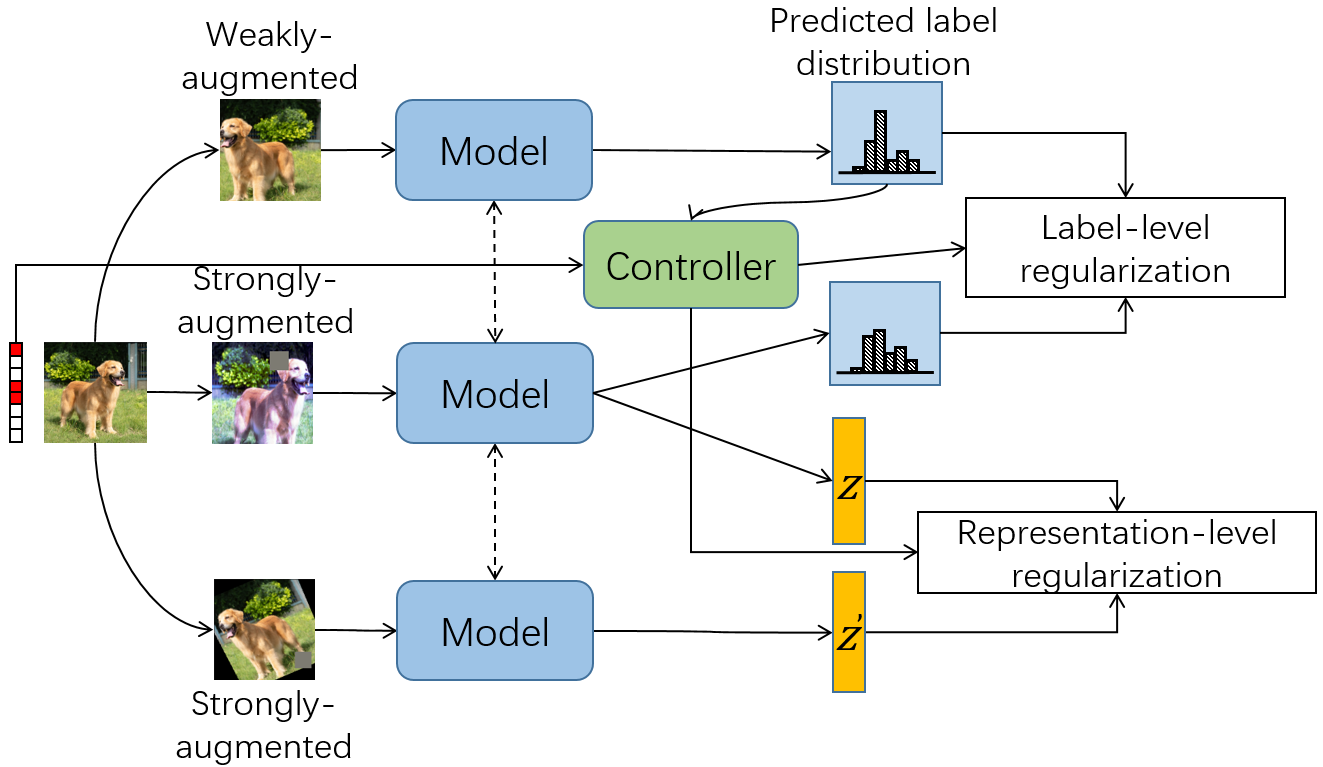}
\caption{The overall framework of ConCont. Dotted lines represent sharing backbone.}
\label{main}
\end{figure}

Let $\tilde{\bm y}^i = f(\textrm{Aug}_w(\bm x^i))$ be the predicted label distribution of the weakly-augmented variant, we predict the pseudo-label of example $(\bm x^i , \bm y^i)$ by selecting the candidate class with the maximum predicted probability: $\hat{y}^i = \arg\max(\tilde{\bm y}^i \circ \bm y^i)$, where $\circ$ means the vector point-wise multiplication. To estimate how confident the model is about the example, the controller computes a partial confidence score (p-score for short) $s^i$ and based on which classifies it as confident or unconfident yet. Given the partial label $\bm y$ and predicted label distribution $\tilde{\bm y}$, the p-score for example $(\bm x , \bm y)$ consists of the following three terms (for the sake of simplicity, we omit the superscript $i$): 

\begin{itemize}

\item Label information: The prior label information from the provided training target. The fewer the number of candidate classes, the greater the amount of information and vice versa. This term is calculated as $p_1 = \frac{1}{\sum_{j=1}^{C} y_j}$.

\item Candidate margin: Measure the prominence in probability between pseudo-label and other candidate classes. We think that the probability margin is more suitable for measuring the confidence than simply the predicted probability over the class. For example, for an instance with 5 candidate classes, the model is obviously more confident in $\tilde{\bm y} = (0.5, 0.2, 0.1, 0.1, 0.1)$, comparing with $\tilde{\bm y}' = (0.5, 0.49, 0.01, 0.0, 0.0)$. This term is calculated as $p_2 = \frac{\max(\tilde{\bm y} \circ \bm y, 1) - \max(\tilde{\bm y} \circ \bm y, 2)}{\sum_{j=1}^{C} \tilde{\bm y}_j \bm y_j}$, where $\max(\cdot, k)$ returns the $k$-th largest value of the input vector.

\item Supervised learning state: The average predicted probability over non-candidate classes, measuring the current learning state of the model from partial supervision, written as $p_3 = \frac{1 - \sum_{j=1}^{C} \tilde{\bm y}_j \bm y_j}{C - \sum_{j=1}^{C} y_j}$

\end{itemize}

The overall p-score is computed as the direct sum of the three terms.
\begin{equation}
p = p_1 + p_2 + p_3.
\end{equation}

\begin{figure}
\subfigure[$q=0.05$]{
\includegraphics[width=3.8cm]{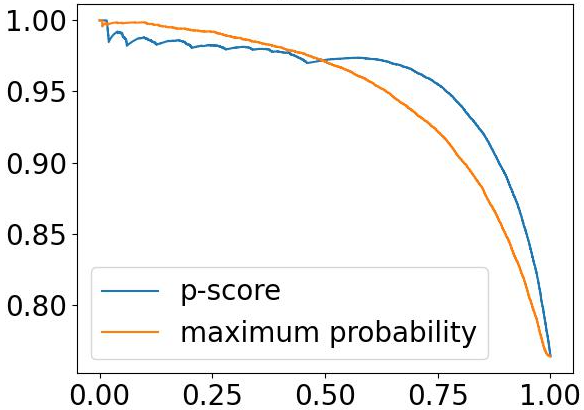}
\label{Hard_PL}
}
\quad
\subfigure[$q=0.1$]{
\includegraphics[width=3.8cm]{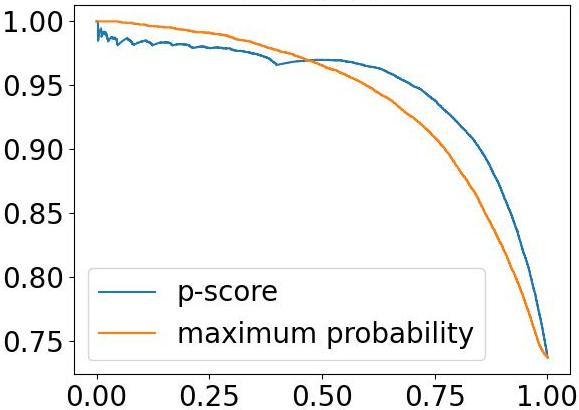}
\label{Hard_pred}
}
\caption{The P-R curves of classifying confident or unconfident examples using p-scores and maximum probabilities, experiments conducted on CIFAR-100 ($40\%$).}
\label{p-score}
\end{figure}

The Precision-Recall (P-R) curves of classifying confident or unconfident examples using p-scores and traditional maximum probabilities are shown in Figure \ref{p-score}. It can be seen that although the curve of p-score has some fluctuations when recall is small, when recall is greater than $60\%$ (in most cases), the precision of p-score is stably higher than the traditional maximum probability.

The label-level regularization is achieved by aligning the output label distributions of strongly-augmented variants with their pseudo-labels for confident samples with p-scores above the threshold. The loss function of example $(\bm x, \bm y)$ can be written as:
\begin{equation}
\mathcal{L}_{\textrm{reg}} = - \mathds{1}(p \geq \tau_{\hat{y}}) \sum_{j=1}^{c} \hat{y}_j \log{f_j(\textrm{Aug}_s(\bm x))}.
\end{equation}
Here, $\hat{y}_j$ is the $j$-th component of the one-hot vector corresponding to the pseudo-label and $\mathds{1}(\cdot)$ returns $1$ if the inside predicate holds, or otherwise returns $0$. $\tau_{\hat{y}}$ is the adaptive threshold for class $\hat{y}$, which will be explained in detail in the next section.

Supervised contrastive learning \cite{DBLP:conf/nips/KhoslaTWSTIMLK20} drawing the projected representations of different augmented variants from the samples of the same class closer and those from different classes apart, is far more effective than simply minimizing the squared loss between two augmented representations and serves as a good way for representation-level consistency regularization. In this paper, since neither partial label nor unlabeled examples have valid labels, we can only proceed with the help of predicted label distributions from $\textrm{Aug}_w(\cdot)$. The controller selects positive and negative samples for each example $(\bm x, \bm y)$ (whose predicted representation denoted as $\bm{z}$) according to the pseudo-labels and p-scores, and the representation distance between positive samples (representation denoted as $\bm{z}_+$) and $\bm x$ are shortened, and the distance between negative samples (representation denoted as $\bm{z}_-$) and $\bm x$ are lengthened. Let $P(\bm x)$ and $N(\bm x)$ separately denote the set of selected positives and negatives, the loss function can be expresses as:
\begin{equation}
\begin{aligned}
\mathcal{L}_{\textrm{reg}}' &= - \frac{1}{|P(\bm x)|} \sum_{\bm{z}_+ \in P(\bm{x})} \\
&\log \frac{\exp(\bm{z}^\top \bm{z}_+ / t)}{\sum\limits_{\bm{z}_+ \in P(\bm{x})} \exp(\bm{z}^\top \bm{z}_+ / t) + \sum\limits_{\bm{z}_- \in N(\bm{x})} \exp(\bm{z}^\top \bm{z}_- / t)},
\end{aligned}
\end{equation}
where $t \geq 0$ is the temperature.

Then, the problem comes to the selection of positive and negative sample. In order to ensure the accuracy of the positive set, at the same time make the negative set as large as possible, which is proven to be crucial by previous studies, our selection strategy is as follows: 1. For a confident input, the confident samples with the same class prediction are selected as positives and all other samples with the different prediction are negatives; 2. For an unconfident input, only the representation of another augmented variant $\textrm{Aug}_s'(\cdot)$ of it is selected as positive and the same with confident input, all sample with different prediction are negatives. To keep the purity of positive set, when facing a sample with the same predicted class as the input, but we are uncertain about one or both of the predictions, the sample is ignored at the current step, i.e., not selected as positive or negative. Since the proportion of such samples is usually lower than $1/C$, discarding them doesn't have a noticeable impact on the total number of contrast representations. In ablation studies, we compare four strategies for sample selection for representation-level consistency regularization and our proposal outperforms the others.

Putting the supervised and the consistency regularized losses together, the overall training objective is: 
\begin{equation}
\mathcal{L} = \mathcal{L}_{\textrm{part}} + \lambda \mathcal{L}_{\textrm{reg}} + \mu \mathcal{L}_{\textrm{reg}}',
\end{equation}
where $\lambda$ and $\mu$ are loss weights.

\begin{table*}[htbp]
\centering
\scalebox{0.8}{
\begin{tabular}{@{}c|c|cccc@{}}
\toprule
Dataset                                                & Method                 & $q=0.1$        & $q=0.3$        & $q=0.5$        & $q=0.7$        \\ \midrule
\multirow{10}{*}{CIFAR-10 (20\%)} & DPLL                   & 88.74$\pm$0.06\%      & 87.15$\pm$0.17\%      & 86.66$\pm$0.16\%      & 63.92$\pm$0.45\%      \\
                                  & DPLL*                  & 90.10$\pm$0.12\% & 89.18$\pm$0.13\% & 82.61$\pm$0.38\% & 68.64$\pm$0.49\%     \\
                                  & PiCO                   & 84.37$\pm$0.34\%      & 83.33$\pm$0.20\%      & 80.67$\pm$0.51\%      & 66.00$\pm$1.22\%      \\
                                  & PiCO*                  & 91.10$\pm$0.18\% & 90.74$\pm$0.50\% & 58.20$\pm$3.78\% & 29.21$\pm$2.47\%     \\
                                  & SSPL                   & 55.67$\pm$0.22\% & 52.51$\pm$0.29\% & 49.62$\pm$0.32\% & 46.10$\pm$0.41\% \\
                                  & PARM                   & 49.12$\pm$0.29\% & 46.04$\pm$0.23\% & 42.84$\pm$0.77\% & 33.13$\pm$3.71\% \\ \cmidrule(l){2-6} 
                                  & DPLL+Controller*       & \textbf{92.95}$\pm$0.03\% 	& \textbf{92.50}$\pm$0.06\%              & 90.82$\pm$0.10\%              & 88.55$\pm$0.15\%             \\
                                  & ConCont (partial only) & 88.45$\pm$0.08\%      & 86.96$\pm$0.07\%      & 85.49$\pm$0.21\%      & 82.73$\pm$0.32\%      \\
                                  & ConCont (two-branch)   & 92.65$\pm$0.03\%      & 92.27$\pm$0.11\%      & 91.44$\pm$0.12\%      & 91.05$\pm$0.15\%      \\
                                  & ConCont                & 92.59$\pm$0.07\%      & 92.22$\pm$0.08\%      & \textbf{91.94}$\pm$0.10\%      & \textbf{91.49}$\pm$0.06\%      \\ \midrule
\multirow{11}{*}{CIFAR-10 (10\%)} & DPLL                   & 81.55$\pm$0.12\%      & 79.20$\pm$0.24\%      & 74.09$\pm$0.37\%      & 28.08$\pm$0.83\%      \\
                                  & DPLL*                  & 85.24$\pm$0.13\% & 80.57$\pm$0.37\% & 71.24$\pm$0.36\% & 15.39$\pm$0.80\%     \\
                                  & PiCO                   & 77.87$\pm$0.44\%      & 76.15$\pm$0.30\%      & 71.01$\pm$0.68\%      & 51.61$\pm$1.69\%      \\
                                  & PiCO*                  & 81.96$\pm$0.53\% & 57.19$\pm$2.88\% & 38.28$\pm$2.02\% & 20.02$\pm$1.36\%     \\
                                  & SSPL                   & 53.17$\pm$0.33\% & 49.75$\pm$0.10\% & 46.79$\pm$0.73\% & 44.10$\pm$0.73\% \\
                                  & PARM                   & 50.19$\pm$0.74\% & 47.15$\pm$0.14\% & 42.16$\pm$1.18\% & 30.69$\pm$2.05\%             \\ \cmidrule(l){2-6} 
                                  & DPLL+Controller*       & 91.00$\pm$0.14\%     & 90.37$\pm$0.21\% & 85.71$\pm$0.36\% & 73.74$\pm$0.85\% 	\\
                                  & ConCont (partial only) & 82.31$\pm$0.17\%      & 80.57$\pm$0.35\%      & 76.69$\pm$0.33\%      & 61.77$\pm$0.69\%      \\
                                  & ConCont (two-branch)   & 90.83$\pm$0.05\%      & 90.39$\pm$0.10\%      & 90.23$\pm$0.09\%      & 81.34$\pm$0.24\%      \\
                                  & ConCont                & \textbf{91.23}$\pm$0.02\%      & \textbf{90.89}$\pm$0.11\% & \textbf{90.46}$\pm$0.16\% & \textbf{82.20}$\pm$0.71\%      \\ \cmidrule(l){2-6} 
                                  & \cellcolor{Gray}Fully Supervised       & \multicolumn{4}{c}{\cellcolor{Gray}94.91$\pm$0.07\%}                          \\ \bottomrule
\end{tabular}}
\caption{Accuracy (mean$\pm$std) comparisons on CIFAR-10 (10\% and 20\%) with different ambiguity levels.}
\label{result1}
\end{table*}

\subsection{Adaptive Confidence Threshold}

PLL often faces severe class-imbalance, especially when there is less partially annotated examples. It is worth noting that the class-imbalance encountered here has different causes from the one we often discuss in supervised learning and long-tailed PLL \cite{wang2022solar, xu2023pseudo, hong2023long}, which is usually caused by the long-tail intrinsic of the dataset. In PLL, the main reason for class-imbalance is that the provided ambiguous supervision allows the model to classify partial examples into any category in the candidate set, and in the consistency regularization lacking of enough correct guidance from initial supervised training model, due to the uneven learning progress of different classes, some strong classes may have so many predictions while the weak classes are gradually overwhelmed.

Inspired by the intrinsic of the problem, we expect to make the consistent regularization process fairer for each class, i.e., make the number of predicted pseudo-labels filtered by the controller for each class as equal as possible. To this end, we propose the adaptive confidence threshold that dynamically adjusts the threshold after each training step for each class, lowering the thresholds for weak classes while raising for strong ones. Specifically, we deem the adaptive thresholds for class $1$ to $C$, denoted as $(\tau_1, \tau_2, \dots, \tau_C)$, as model parameters and minimize the cross-entropy between the uniform distribution and the distribution of the numbers of filtered pseudo-labels via gradient descent. 

Let $s_j = Count_j(\tau_j)$ denote the count number of pseudo-labels of class $j$ with p-scores above threshold $\tau_j$ at the current training step, the distribution of the numbers of filtered pseudo-labels can be expressed as: $P = (s_1/s', s_2/s', \dots, s_C/s')$, where $s' = \sum_{j=1}^{C} s_j$ is the summation of all filtered pseudo-labels, and the uniform distribution is written as: $U = (1/C, 1/C, \dots, 1/C)$. 

The partial derivative of the cross-entropy loss $H(U; P)$ with respect to threshold $\tau_j$ is calculated as:
\begin{equation}
\label{derivation}
\begin{aligned}
\frac{\partial H(U; P)}{\partial \tau_j} & = \frac{\partial H(U; P)}{\partial s_j} \cdot \frac{\partial s_j}{\partial \tau_j}\\
& = \frac{1}{\ln{2}} \cdot \frac{s_j - \bar{s}}{s' \cdot s_j} \cdot \frac{\partial s_j}{\partial \tau_j},
\end{aligned}
\end{equation}
where $\bar{s} = s'/C$ is the mean value of the number of filtered pseudo-labels. The detailed derivation of Eq. \ref{derivation} can be found in Appendix.

The last term of Eq. \ref{derivation} can also be written as $Count_j'(\tau_j)$. The $Count_j(\cdot)$ function decreases monotonically from the number of all predictions of class $j$ to $0$ as the  input threshold increases, and $Count_j'(\tau_j)$ reflects the sample density where the p-score is in the neighborhood of $\tau_j$. Since samples are fed into the model in batches during training and the number of samples of each class in each batch is too small to estimate $Count_j'(\tau_j)$. In actual implementation, we think that the density of samples can be considered to be roughly proportional to the number of filtered pseudo-labels $s_j$. The update of $\tau_j$ through gradient descent turns to be:
\begin{equation}
\tau_j = \tau_j - \frac{\bar{s} - s_j}{s'} \cdot \gamma_\tau,
\end{equation}
where $\gamma_\tau$ is the learning rate for adaptive thresholds.

Finally, the thresholds are clamped to a preset interval.
\begin{equation}
\tau_j = \min(\max(\tau_j, \tau_{l}), \tau_{u}),
\end{equation}
where $\tau_{l}$ and $\tau_{u}$ denote the lower and upper bound of adaptive thresholds, respectively.

\begin{table*}[htbp]
\centering
\scalebox{0.8}{
\begin{tabular}{@{}c|c|cccc@{}}
\toprule
Dataset                                                & Method                 & $q=0.1$        & $q=0.3$        & $q=0.5$        & $q=0.7$        \\ \midrule
\multirow{10}{*}{SVHN (20\%)}     & DPLL                   & 95.84$\pm$0.20\%      & 95.00$\pm$0.21\%      & 94.30$\pm$0.02\%      & 59.50$\pm$1.82\%      \\
                                  & DPLL*                  & 90.84$\pm$0.24\% & 83.25$\pm$0.38\% & 62.37$\pm$2.21\% & 24.47$\pm$1.83\% \\
                                  & PiCO                   & 94.35$\pm$0.19\%      & 42.66$\pm$1.16\%      & 20.44$\pm$0.49\%      & 13.86$\pm$0.44\%      \\
                                  & PiCO*                  & 71.93$\pm$0.35\% & 30.29$\pm$1.06\% & 16.79$\pm$0.94\% & 11.93$\pm$0.59\% \\
                                  & SSPL                   & 75.43$\pm$0.09\% & 73.93$\pm$0.15\% & 71.96$\pm$0.12\% & 69.00$\pm$0.32\% \\
                                  & PARM                   & 70.80$\pm$0.57\% & 68.28$\pm$0.61\% & 63.80$\pm$2.26\% & 61.87$\pm$0.66\% \\ \cmidrule(l){2-6} 
                                  & DPLL+Controller*       & 96.98$\pm$0.18\%      & 96.47$\pm$0.09\%      & 94.88$\pm$0.11\%      & 82.39$\pm$0.22\%      \\
                                  & ConCont (partial only) & 96.40$\pm$0.09\%      & 96.39$\pm$0.09\%      & \textbf{95.96}$\pm$0.04\%      & 80.51$\pm$0.31\%      \\
                                  & ConCont (two-branch)   & 97.05$\pm$0.10\%      & \textbf{96.98}$\pm$0.02\%      & 94.81$\pm$0.09\%      & 90.92$\pm$0.10\%      \\
                                  & ConCont                & \textbf{97.07}$\pm$0.14\%      & 96.92$\pm$0.12\%      & \textbf{95.68}$\pm$0.05\%      & \textbf{94.25}$\pm$0.07\%      \\ \midrule
\multirow{11}{*}{SVHN (10\%)}     & DPLL                   & 94.64$\pm$0.04\%      & 94.37$\pm$0.14\%      & 90.30$\pm$0.13\%      & 37.79$\pm$2.37\%      \\
                                  & DPLL*                  & 79.15$\pm$0.64\% & 55.06$\pm$1.69\% & 19.93$\pm$1.02\% & 19.65$\pm$0.80\% \\
                                  & PiCO                   & 89.34$\pm$0.16\%      & 60.06$\pm$1.00\%      & 16.48$\pm$0.24\%      & 13.86$\pm$0.33\%      \\
                                  & PiCO*                  & 37.60$\pm$1.98\% & 34.30$\pm$2.09\% & 14.55$\pm$1.34\% & 13.25$\pm$0.81\% \\
                                  & SSPL                   & 73.54$\pm$0.04\% & 71.58$\pm$0.14\% & 69.21$\pm$0.34\% & 65.51$\pm$0.42\% \\
                                  & PARM                   & 67.94$\pm$0.96\% & 66.95$\pm$1.36\% & 62.69$\pm$2.73\% & 59.12$\pm$0.71\% \\ \cmidrule(l){2-6} 
                                  & DPLL+Controller*       & 96.33$\pm$0.09\%      & 95.40$\pm$0.16\%      & 92.89$\pm$0.23\%      & 70.02$\pm$1.45\%      \\
                                  & ConCont (partial only) & 95.12$\pm$0.08\%      & 95.04$\pm$0.06\%      & 94.25$\pm$0.12\%      & 87.67$\pm$0.09\%      \\
                                  & ConCont (two-branch)   & \textbf{96.72}$\pm$0.01\%      & 96.19$\pm$0.02\%      & 93.76$\pm$0.13\%      & 88.59$\pm$0.18\%      \\
                                  & ConCont                & 96.62$\pm$0.06\%      & \textbf{96.40}$\pm$0.10\%      & \textbf{94.40}$\pm$0.07\%      & \textbf{88.71}$\pm$0.07\%      \\ \cmidrule(l){2-6} 
                                  & \cellcolor{Gray}Fully Supervised       & \multicolumn{4}{c}{\cellcolor{Gray}97.35\%$\pm$0.12\%}                        \\ \bottomrule
\end{tabular}}
\caption{Accuracy (mean$\pm$std) comparisons on SVHN (10\% and 20\%) with different ambiguity levels.}
\label{result2}
\end{table*}

\section{Experiments}
\subsection{Experimental Setup}

We evaluate on commonly used benchmarks CIFAR-10 \cite{krizhevsky2009learning} and SVHN \cite{goodfellow2013multi} with $10\%$ and $20\%$ partial examples, and CIFAR-100 \cite{krizhevsky2009learning} with $20\%$ and $40\%$ partial examples. To synthesize partial labels, negative labels are randomly flipped to false positive ones with a probability $q$. The comparing methods include state-of-the-art PLL methods DPLL \cite{wu2022revisiting} and PiCO \cite{wang2022pico} (* indicates the version using unlabeled examples), as well as traditional semi-supervised PLL methods SSPL \cite{wang2019partial} and PARM \cite{wang2020semi}. ConCont (partial only) shows the results of our method without unlabeled examples, and ConCont (two-branch) is the previously mentioned "two-branch" implementation. To examine the versatility of our method, we additionally adapt the controller module (as well as adaptive thresholds) to DPLL, denoted as DPLL+Controller*. We record the mean and standard deviation based on three independent runs. Additional details of the experiments can be found in the Appendix.

\subsection{Main Results}

Table \ref{result1}, \ref{result2} and \ref{result3} report the main experimental results. As is shown, our method consistently outperforms all previous methods. For example, on CIFAR-10 dataset with $10\%$ partial examples, our method improves upon the best baseline by $5.99\%$, $10.32\%$ and $16.37\%$ where $q$ is set to $0.1$, $0.3$ and $0.5$ respectively. We can also observe that our method achieves decent performances despite some degradation in some extreme cases with rare partial annotation and high ambiguity, where previous methods fail to obtain usable results, which extends the practical applicability of PLL. It is also noteworthy that adapting the controller and adaptive thresholds to DPLL achieves satisfactory results, which shows that the controller inspires the capability of integrating unlabeled examples of DPLL, demonstrating the versatility of our components. 

\subsection{Ablation Studies}

\begin{table}[htbp]
\centering
\scalebox{0.9}{
\begin{tabular}{c|cc}
\toprule
Ablation	& \makecell{CIFAR-10 \\ (10\%, $q=0.5$)}	& \makecell{CIFAR-100 \\ (20\%, $q=0.1$)} \\ \midrule
ConCont	& \textbf{90.46}$\pm$0.16\%	& \textbf{62.85}$\pm$0.26\% \\
w/o $\mathcal{L}_{\textrm{part}}$ 	& 44.32$\pm$1.51\%	& 18.84$\pm$0.87\% \\ 
w/o $\mathcal{L}_{\textrm{reg}}$	& 80.62$\pm$0.34\%	& 55.91$\pm$0.35\% \\
w/o $\mathcal{L}_{\textrm{reg}}'$	& 86.76$\pm$0.13\%	& 60.05$\pm$0.19\% \\
w/o $\mathcal{L}_{\textrm{reg}}$ and $\mathcal{L}_{\textrm{reg}}'$ 	& 70.93$\pm$0.21\%	& 49.42$\pm$0.37\% \\ \bottomrule
\end{tabular}}
\caption{Ablation study on different components of our method.}
\label{ab}
\end{table}

\subsubsection{Effects of supervised loss and consistency regularization.}

We ablate the contributions of the supervised loss and consistency regularization. Specifically, we compare our method with four degenerations: 1. w/o $\mathcal{L}_{\textrm{part}}$ substitutes the partial cross-entropy loss with the traditional cross-entropy loss for regular supervised learning; 2. w/o $\mathcal{L}_{\textrm{reg}}$ does not perform label-level consistency regularization; 3. w/o $\mathcal{L}_{\textrm{reg}}'$ does not perform representation-level consistency regularization; w/o $\mathcal{L}_{\textrm{reg}}$ and $\mathcal{L}_{\textrm{reg}}'$ does not perform consistency regularization at both levels. Table \ref{ab} reports the accuracy comparisons on $10\%$ partially annotated CIFAR-10 dataset with $q=0.5$ and $20\%$ partially annotated CIFAR-100 dataset with $q=0.1$. The bad results achieved by w/o $\mathcal{L}_{\textrm{part}}$ indicates that a supervised loss that conforms to the partial label semantics is fundamental. It can also be seen that the consistency regularization at both levels contribute significantly the model performance.

\begin{table}[htbp]
\centering
\scalebox{0.9}{
\begin{tabular}{@{}c|cc@{}}
\toprule
Method	& \makecell{CIFAR-10 \\ (10\%, $q=0.5$)}	& \makecell{CIFAR-100 \\ (20\%, $q=0.1$)} \\ \midrule
Adaptive	& \textbf{90.46}$\pm$0.16\% 	& \textbf{62.85}$\pm$0.26\%   \\
DA		& 89.08$\pm$0.25\% 	& 61.04$\pm$0.02\%   \\
CPL		& 88.83$\pm$0.14\%	& 46.32$\pm$1.21\%   \\
Vanilla		& 86.50$\pm$0.21\%  	& 59.50$\pm$0.28\%   \\ \midrule
Adaptive	& 90.32$\pm$0.12\% 	& \textbf{62.86}$\pm$0.23\%   \\
DA (soft)	& 88.77$\pm$0.15\%  	& 61.07$\pm$0.13\%   \\
Vanilla (soft)  	& 87.95$\pm$0.01\% 	& 60.10$\pm$0.25\%   \\ \bottomrule
\end{tabular}}
\caption{Accuracy comparisons of different class-imbalance methods. Vanilla means no additional operation alleviating class-imbalance, and "(soft)" means the usage of KL-divergence of label distribution instead of the cross-entropy loss with one-hot pseudo-label in the label-level consistency regularization.}
\label{ab_pl}
\end{table}

\begin{table*}[htbp]
\centering
\scalebox{0.8}{
\begin{tabular}{@{}c|c|cccc@{}}
\toprule
Dataset                            & Method                 & $q=0.05$       & $q=0.1$        & $q=0.15$       & $q=0.2$        \\ \midrule
\multirow{10}{*}{CIFAR-100 (40\%)} & DPLL                   & 68.07$\pm$0.13\%      & 66.51$\pm$0.10\%      & 62.37$\pm$0.08\%      & 55.54$\pm$0.37\%      \\
                                   & DPLL*                  & 68.02$\pm$0.26\% & 64.92$\pm$0.27\% & 54.78$\pm$0.56\%     & 45.60$\pm$2.25\% \\
                                   & PiCO                   & 57.62$\pm$0.15\%      & 40.07$\pm$0.28\%      & 26.04$\pm$0.69\%      & 17.32$\pm$0.47\%      \\
                                   & PiCO*                  & 64.01$\pm$0.35\% & 40.06$\pm$1.82\% & 21.52$\pm$1.22\%     & 18.61$\pm$0.56\% \\
                                   & SSPL                   & 22.20$\pm$0.10\% & 20.33$\pm$0.13\% & 18.95$\pm$0.06\% & 17.94$\pm$0.04\% \\
                                   & PARM                   & /            & /            & /            & /            \\ \cmidrule(l){2-6} 
                                   & DPLL+Controller*       & \textbf{70.17}$\pm$0.20\%      & 68.10$\pm$0.23\%      & 64.33$\pm$0.18\%      & 58.02$\pm$0.37\%      \\
                                   & ConCont (partial only) & 66.24$\pm$0.10\%      & 64.02$\pm$0.13\%      & 60.96$\pm$0.26\%      & 56.03$\pm$0.37\%      \\
                                   & ConCont (two-branch)   & 68.95$\pm$0.10\%      & 67.86$\pm$0.18\%      & \textbf{66.03}$\pm$0.22\%      & 59.41$\pm$0.27\%      \\
                                   & ConCont                & 69.17$\pm$0.29\%      & \textbf{68.46}$\pm$0.16\%      & 65.70$\pm$0.21\%      & \textbf{60.59}$\pm$0.18\%      \\ \midrule
\multirow{11}{*}{CIFAR-100 (20\%)} & DPLL                   & 58.59$\pm$0.27\%      & 51.61$\pm$0.17\%      & 39.56$\pm$0.44\%      & 28.21$\pm$0.23\%      \\
                                   & DPLL*                  & 57.25$\pm$0.28\% & 41.03$\pm$0.49\% & 29.36$\pm$0.72\%     & 24.71$\pm$0.45\% \\
                                   & PiCO                   & 49.05$\pm$0.64\%      & 27.67$\pm$0.95\%      & 14.85$\pm$0.40\%      & 9.59$\pm$0.83\%       \\
                                   & PiCO*                  & 47.17$\pm$0.73\% & 27.10$\pm$1.51\% & 17.70$\pm$1.44\%     & 9.65$\pm$1.69\%  \\
                                   & SSPL                   & 18.73$\pm$0.23\% & 16.67$\pm$0.18\% & 15.61$\pm$0.10\% & 14.24$\pm$0.35\% \\
                                   & PARM                   & /            & /            & /            & /            \\ \cmidrule(l){2-6} 
                                   & DPLL+Controller*       & \textbf{65.20}$\pm$0.04\%      & 58.02$\pm$0.17\% & 51.89$\pm$0.58\%      & 40.42$\pm$1.12\%      \\
                                   & ConCont (partial only) & 56.43$\pm$0.12\%      & 51.63$\pm$0.04\%      & 44.32$\pm$0.08\%      & 38.60$\pm$0.28\%      \\
                                   & ConCont (two-branch)   & 63.86$\pm$0.05\%      & 62.00$\pm$0.27\%      & 58.07$\pm$0.14\%      & 47.30$\pm$0.16\%      \\
                                   & ConCont                & 63.55$\pm$0.10\%      & \textbf{62.85}$\pm$0.26\% & \textbf{58.62}$\pm$0.02\%      & \textbf{49.42}$\pm$0.31\%      \\ \cmidrule(l){2-6} 
                                   & \cellcolor{Gray}Fully Supervised       & \multicolumn{4}{c}{\cellcolor{Gray}73.56$\pm$0.10\%}                          \\ \bottomrule
\end{tabular}}
\caption{Accuracy (mean$\pm$std) comparisons on CIFAR-100 (20\% and 40\%) with different ambiguity levels. "/" means not applicable.}
\label{result3}
\end{table*}

\subsubsection{Comparison of different class-imbalance methods.}

To demonstrate the superiority of the adaptive confidence threshold, we further experiment with two widespread class-imbalance methods working on pseudo-labels, i.e., Distribution Alignment (DA) from ReMixMatch \cite{berthelot2019remixmatch} and Curriculum Pseudo-Labeling (CPL) from FlexMatch \cite{zhang2021flexmatch} . It can be seen from Table \ref{ab_pl} that the performance improvements brought by alleviating class-imbalance are significant. Our method brings more than $3\%$ accuracy improvements on both datasets compared to not using and is better than other class-imbalance methods (outperforms the best comparing method by $1.38\%$ on CIFAR-10 (10\%, $q=0.5$) and $1.81\%$ on CIFAR-100 (20\%, $q=0.1$)).

Additionally, we explore the usage of KL-divergence of label distribution instead of the cross-entropy loss with one-hot pseudo-labels in the label-level consistency regularization, i.e., soft pseudo-labels. The same with one-hot pseudo-labels, we compute p-scores and mask out unconfident examples via adaptive thresholds. As shown in Table \ref{ab_pl}, our method still achieves the best performances. CPL is not compared here since it is designed for one-hot pseudo-labels.

\begin{table}[htbp]
\centering
\scalebox{0.9}{
\begin{tabular}{c|cc}
\toprule
Strategy	& \makecell{CIFAR-10 \\ (10\%, $q=0.5$)}		& \makecell{CIFAR-100 \\ (20\%, $q=0.1$)}    \\ \midrule
Ours		& \textbf{90.46}$\pm$0.16\%		& \textbf{62.85}$\pm$0.26\% \\
HCPN			& 89.25$\pm$0.16\%		& 62.50$\pm$0.17\% \\
HCP			& 89.76$\pm$0.30\%		& 62.35$\pm$0.27\% \\
SupCon		& 89.32$\pm$0.08\%		& 62.22$\pm$0.11\% \\
UnsupCon		& 89.70$\pm$0.01\%		& 61.68$\pm$0.31\% \\ \midrule
w/o $\mathcal{L}_{\textrm{reg}}'$	& 86.76$\pm$0.13\%	& 60.05$\pm$0.19\% \\
Y\_true & 90.66$\pm$0.09\%	& 64.50$\pm$1.16\% \\ \bottomrule
\end{tabular}}
\caption{Accuracy comparisons among different strategies.}
\label{ab_cont}
\end{table}

\subsubsection{Different selection strategies of positive and negative samples.}

In order to ensure the accuracy of the positive set, at the same time make the negative set with high error-tolerance as large as possible, our method includes a tailor-made sample selection strategy in the representation-level consistency regularization. Here, we compare our method with other strategies including: UnsupCon, SupCon, HCP and HCPN, their specific ways are listed as follows:
\begin{itemize}

\item UnsupCon: Traditional unsupervised contrastive learning, selects the representation of $\textrm{Aug}_s'(\cdot)$ as the only positive sample.

\item SupCon: Directly use pseudo-labels as ground-truth and performs supervised contrastive learning.

\item High-Confidence Positive (HCP): Ensure the accuracy of positive set and put all remaining samples into the negative set. For a confident input, select confident samples with the same prediction as positives; while for an unconfident input, only select the representation of its another view as positive.

\item High-Confidence Positive and Negative (HCPN): Ensure the accuracy of both sets. For a confident input, we separately select confident samples with or without the same pseudo-label as positives or negatives, while neglecting the unconfident input when computing $\mathcal{L}_{\textrm{reg}}'$.

\end{itemize}
In order to reflect the differences more clearly, we use true unique-labels as the supervision for supervised contrastive learning serving as the upper bound (denoted as Y\_true), and w/o $\mathcal{L}_{\textrm{reg}}'$ as the lower bound.

As shown in table \ref{ab_cont}, our method performs the best among all comparing strategies on both cases. On CIFAR-10 (10\%, $q=0.5$), our method outperforms the second-ranked HCP by $0.7\%$ and is only $0.2\%$ below the upper bound. On CIFAR-100 (20\%, $q=0.1$), HCPN performs the best among other comparing strategies (mainly because that it only uses high-confident samples and ignores uncertain ones), and our method outperforms HCPN by $0.35\%$.

\section{Conclusion}
In this paper, we observe that the current mainstream, consistent regularization-based PLL methods are flawed when there are not enough partially labeled examples, which greatly limits their practical usability. To address this issue, we introduce cheaply acquired unlabeled examples and devise a method for consistency regularization at the label and representation levels. The highlights of our approach are the p-scores computed by the controller and how they are used to control the subsequent consistency regularization, as well as the simple yet effective class-balancing mechanism. Experiments show that our method can achieve satisfactory results with only a small number of partial examples and unlabeled examples.

\section{Acknowledgments}
This work is supported in part by the National Natural Science Foundation of China under Grant 62171248, Shenzhen Science and Technology Program (JCYJ20220818101012025), and the PCNL KEY project (PCL2023AS6-1).

\bibliography{aaai24}

\appendix
\newpage

\section{Derivation of Eq. (6)}

Let $s_j$ denote the number of filtered samples with pseudo-label $j$, the distribution of the proportion of filtered pseudo-labels of each class is written as $P = (s_1/s', s_2/s', \dots, s_C/s')$, where $s' = \sum_{j=1}^{C} s_j$ is the summation of $s_j$, and the uniform distribution is written as: $U = (1/C, 1/C, \dots, 1/C)$. The cross-entropy loss between $U$ and $P$ can be derived as follows:
\begin{equation}
\begin{aligned}
H(U; P) &= - \sum_{k=1}^{C} \frac{1}{C} \cdot \log \frac{s_k}{s'} \\
&= - \frac{1}{C} \cdot [\log \frac{s_j}{s'} + \sum_{k \neq j} \log \frac{s_k}{s'}] \\
&= - \frac{1}{C} \cdot [\log s_j - \log s' + \sum_{k \neq j} (\log s_k - \log s')] \\
&= - \frac{1}{C} \cdot \log s_j + \log s' - \frac{1}{C} \cdot \sum_{k \neq j} s_k.
\end{aligned}
\end{equation}

Calculate the partial derivative of $H(U; P)$ with respect to $s_j$, we have:
\begin{equation}
\begin{aligned}
\frac{\partial H(U; P)}{\partial s_j} &= - \frac{1}{C} \cdot  \frac{1}{\ln2 \cdot s_j} + \frac{1}{\ln2 \cdot s'} \\
&= - \frac{1}{\ln2} \cdot \frac{s_j - \bar{s}}{s' \cdot s_j},
\end{aligned}
\end{equation}
where $\bar{s} = s'/C$ is the mean value of the number of filtered pseudo-labels. Substituting the above into Eq. (6), the derivation is completed.

\section{Practical Implementations}
\subsection{Data Augmentations}
We use two types of data augmentations in our method. The weak augmentation is implemented by randomly flipping the images horizontally with a probability of $50\%$, and randomly cropping the images with a scale interval of $(0.5, 1)$ and then resizing to the normal size. We implement the strong augmentation by first performing random horizontal flipping with a probability of $50\%$ and random resized cropping with a scale interval of $(0.2, 1)$. Then, we randomly select from the augmentation policies released by AutoAugment \cite{cubuk2019autoaugment} to perform, i.e.,  RandAugment \cite{DBLP:journals/corr/abs-1909-13719}, and lastly apply Cutout \cite{devries2017improved}.

\subsection{Momentum constrast}
\begin{figure}[t]
\centering
\includegraphics[scale=0.22]{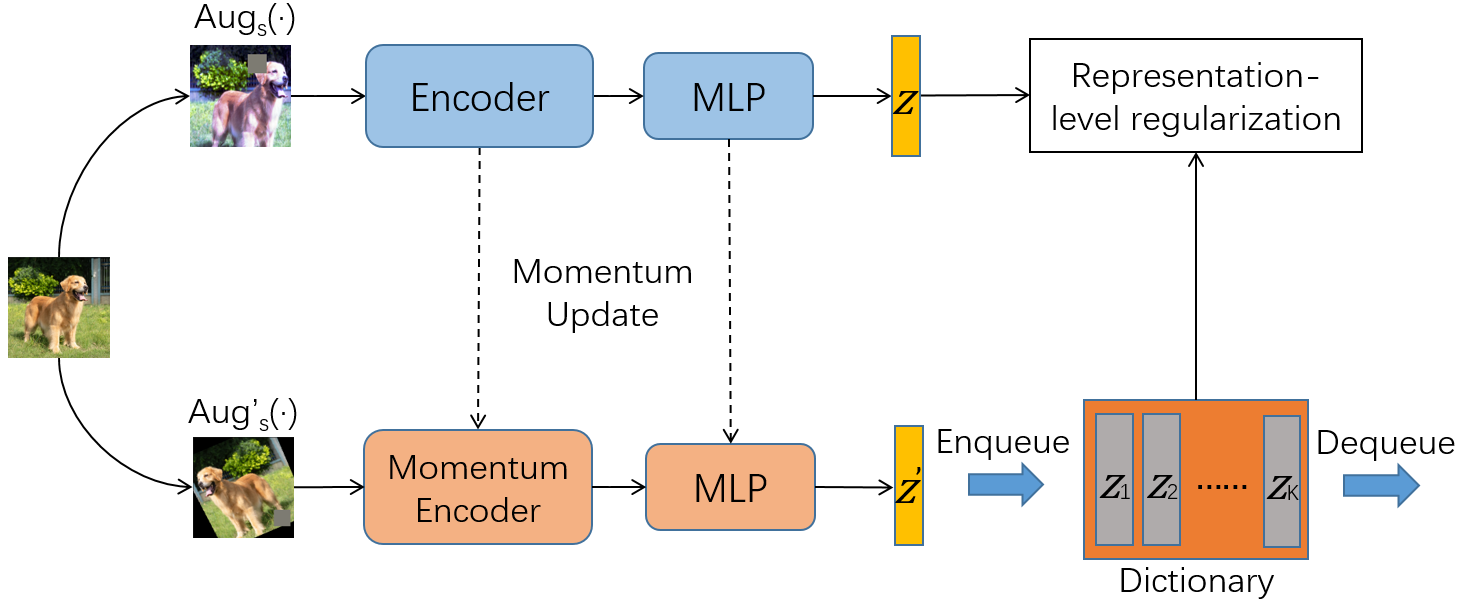}
\caption{Momentum constrast.}
\label{main}
\end{figure}

To expand the number of contrast representations and enhance their consistency, our concrete framework is implemented in terms of MoCo \cite{DBLP:conf/cvpr/He0WXG20}. As shown in Figure \ref{main}, given each training image $(\bm x, \bm y)$, our method generates two views by imposing strong data augmentations $\textrm{Aug}_s(\cdot)$ and $\textrm{Aug}_s'(\cdot)$. Then, the two views are fed separately into the main encoder $g(\cdot)$ and the momentum encoder $g'(\cdot)$, passing through MLP projection heads $\textrm{Proj}(\cdot)$ and $\textrm{Proj}'(\cdot)$, outputting a pair of normalized representation $\bm z = \textrm{Proj}(g(\textrm{Aug}_s(\bm x)))$ and $\bm z' = \textrm{Proj}'(g'(\textrm{Aug}_s'(\bm x)))$, called query and key representations in the original paper \cite{DBLP:conf/cvpr/He0WXG20}. The key representations form a dictionary maintained as a queue (first-in-first-out) and are matched by query representations to perform the representation-level regularization.

\section{Additional Experimental Setups}
\subsection{Datasets}

We evaluate our method on three commonly used benchmark image datasets: CIFAR-10, CIFAR-100 \cite{krizhevsky2009learning} and SVHN \cite{goodfellow2013multi}. To experiment with limited partial annotations, we randomly sample $10\%$ and $20\%$ training images for CIFAR-10 and SVHN, and $20\%$ and $40\%$ for CIFAR-100 as partially labeled examples, and the remaining are unlabeled. We then generate partial labels by flipping negative labels $\hat{\ell} \neq \ell$ to false positive ones with a probability $q = P(y_{\hat{\ell}}=1 | \hat{\ell} \neq \ell)$, i.e. ambiguity level, and the flipped ones are aggregated with the ground-truth labels to form the candidate sets, which the same with the previous works \cite{DBLP:conf/icml/LvXF0GS20, wang2022pico}. In this paper, we consider $q \in \{0.1, 0.3, 0.5, 0.7\}$ for CIFAR-10 and SVHN and $q \in \{0.05, 0.1, 0.15, 0.2\}$ for CIFAR-100.

\subsection{Comparing methods}

We compare our method with two best-performed PLL methods: 1. DPLL \cite{wu2022revisiting} is a novel method which performs supervised learning on non-candidate labels and consistency regularization over the label distribution of multiple augmented variants on candidate labels. 2. PiCO \cite{wang2022pico} adopts the contrastive learning technique for representation alignment and identifies ground-truth labels with learned prototype embeddings. We implement two versions for DPLL and PiCO, one using only partial examples, and one using both partial and unlabeled examples (add '*' behind the methods), where unlabeled examples are treated like partial examples whose candidate set equals to the entire label space. As well as two representative semi-supervised PLL methods: 3. SSPL \cite{wang2019partial}, in which partial label disambiguation and unlabeled exploitation is conducted simultaneously with an iterative label propagation procedure via weighted graph. 4. PARM \cite{wang2020semi} maximizes the confidence-rated margin by preserving labeling confidence manifold structure over partial label and unlabeled data.

We experiment with four variants/degradations of our method: 1. ConCont is the original proposed method. 2. ConCont (two-branch) is the previously mentioned two-branch version, whose label- and representation-level consistency regularization is performed on the same pair of augmentations $(\textrm{Aug}_w(\cdot), \textrm{Aug}_s(\cdot))$, while the complete method perform label-level on $(\textrm{Aug}_w(\cdot), \textrm{Aug}_s(\cdot))$ and representation-level on $(\textrm{Aug}_s(\cdot), \textrm{Aug}_s'(\cdot))$. 3. ConCont (partial only) shows the performance of our method using only partially labeled examples, demonstrating the competitive results of our method as a pure PLL approach and the performance gain obtained by introducing unlabeled examples. 4. DPLL+Controller* is the method adapting our consistency controller as well as adaptive thresholds to DPLL. The controller computes p-scores for conformal label distribution to decide whether to use for consistency regularization. The convincing results it achieved show that the adapted controller inspires DPLL's ability of integrating unlabeled examples for better consistency regularization and the versatility of the proposed module.

To make fair comparisons, we employ the same backbone network, i.e. ResNet-18 for all deep methods. And for the others, we extract the 512-dimensional GIST \cite{oliva2001modeling} feature for image representation. We basically use the hyper-parameters specified in their original papers, except on some particular cases, the hyper-parameters of comparing methods are further fine-tuned. PARM needs to perform quadratic programming with a $C^2$-by-$C^2$ dimensional quadratic coefficient matrix for $p+u$ times in each optimization iteration, which cannot be completed in tolerable time on CIFAR-100.

\subsection{Hyper-parameters}

The implementation of our method is based on PyTorch \cite{paszke2019pytorch}, and experiments were carried out with NVIDIA series GPUs. The loss weights of overall training objective are set as $\lambda=1$ and $\mu=0.1$ respectively. The initial value, lower and upper bound of adaptive thresholds are set as $0.8, [0.5, 0.95]$ for CIFAR-10 and SVHN, and $0.6, [0.4, 0.8]$ for CIFAR-100, and its learning rate is set as $\gamma_\tau=1.0$. The projection head of contrastive learning is implemented with a 2-layer MLP outputting 64-dimensional representations. The size of the dictionary storing representations is fixed to $8192$, and the coefficient for momentum network updating is set as $0.999$. We use SGD as the optimizer with a momentum of $0.9$ and a weight decay of $1e-3$. The batch size is set to $256$. We train the model for 800 total epochs, and the initial learning rate is set as $0.01$ with cosine learning rate scheduling.

\end{document}